\documentclass[runningheads]{llncs}

\usepackage{eccv}

\usepackage{eccvabbrv}

\usepackage{graphicx}
\usepackage{booktabs}

\usepackage[accsupp]{axessibility}  % Improves PDF readability for those with disabilities.

\usepackage{multirow}
\usepackage{placeins}
\usepackage[table]{xcolor}

\newcommand{\simobject}{\mbox{sim-object}\xspace}
\newcommand{\simhuman}{\mbox{sim-human}\xspace}
\newcommand{\gr}{\cellcolor{gray!15}--} 

\usepackage{hyperref}
\usepackage{marvosym}

\usepackage{orcidlink}

\begin{document}

% ---------------------------------------------------------------
% TODO REVIEW: Replace with your title
\title{WiFi-JEPA: Self-supervised Learning for WiFi-CSI 3D Human Pose Estimation} 

% TODO REVIEW: If the paper title is too long for the running head, you can set
% an abbreviated paper title here. If not, comment out.
\titlerunning{WiFi-JEPA}

% TODO FINAL: Replace with your author list. 
% Include the authors' OCRID for the camera-ready version, if at all possible.
\author{
Doeon Kim\inst{1}\thanks{These authors contributed equally to this work.}\orcidlink{0009-0006-3187-4503} \and
Jungyoon Lee\inst{2}\textsuperscript{$\star$}\orcidlink{0009-0001-0240-4741} \and
Seongsin Kim\inst{3}\textsuperscript{\Letter}\orcidlink{0009-0004-0260-6257} \and
Seong-heum Kim\inst{1}\orcidlink{0000-0003-2551-0157}
}

% TODO FINAL: Replace with an abbreviated list of authors.
\authorrunning{D.~Kim, J.~Lee et al.}
% First names are abbreviated in the running head.
% If there are more than two authors, 'et al.' is used.

% TODO FINAL: Replace with your institution list.
\institute{Department of Intelligent Semiconductors, Soongsil University, Republic of Korea\and
Department of AI Convergence Security, Soongsil University, Republic of Korea\and
School of AI Software, Soongsil University, Republic of Korea\\
\email{\{ilsin205, jungyoon\}@soongsil.ac.kr}, \email{\{kss0222, seongheum\}@ssu.ac.kr}\\
}

\maketitle

\begin{abstract}
 WiFi Channel State Information (CSI) enables \linebreak privacy-preserving human pose sensing in camera-denied environments, but existing WiFi-based pose estimators often fail under environment shifts and rely on costly camera-based annotation pipelines that limit scale. We propose \textbf{WiFi-JEPA}, a self-supervised framework that learns CSI-native representations by predicting masked latent embeddings instead of reconstructing raw CSI signals that may contain hardware-specific artifacts. WiFi-JEPA makes three contributions: (i) CSI-specific tokenization and link masking tailored to the CSI tensor over channel, time, and link $(C,T,L)$; masking entire Tx–Rx antenna links forces the model to predict one spatial link view from others, capturing cross-link correlations informative of 3D spatial structure. (ii) A ray-tracing CSI simulation pipeline that generates diverse unlabeled CSI from randomized geometric primitives, providing scalable pre-training data without pose annotations. (iii) State-of-the-art results on Person-in-WiFi-3D: WiFi-JEPA outperforms prior WiFi-CSI baselines on both single- and multi-person 3D pose estimation under the same evaluation protocol. We also show that simulated CSI provides complementary pre-training signal to real CSI, and that four vision-native SSL objectives degrade performance below training from scratch, whereas WiFi-JEPA consistently improves downstream pose estimation.
 
\end{abstract}

\begin{figure}[!t]
  \centering
  \includegraphics[trim=0cm 1cm 0cm 0cm, width=0.9\linewidth]{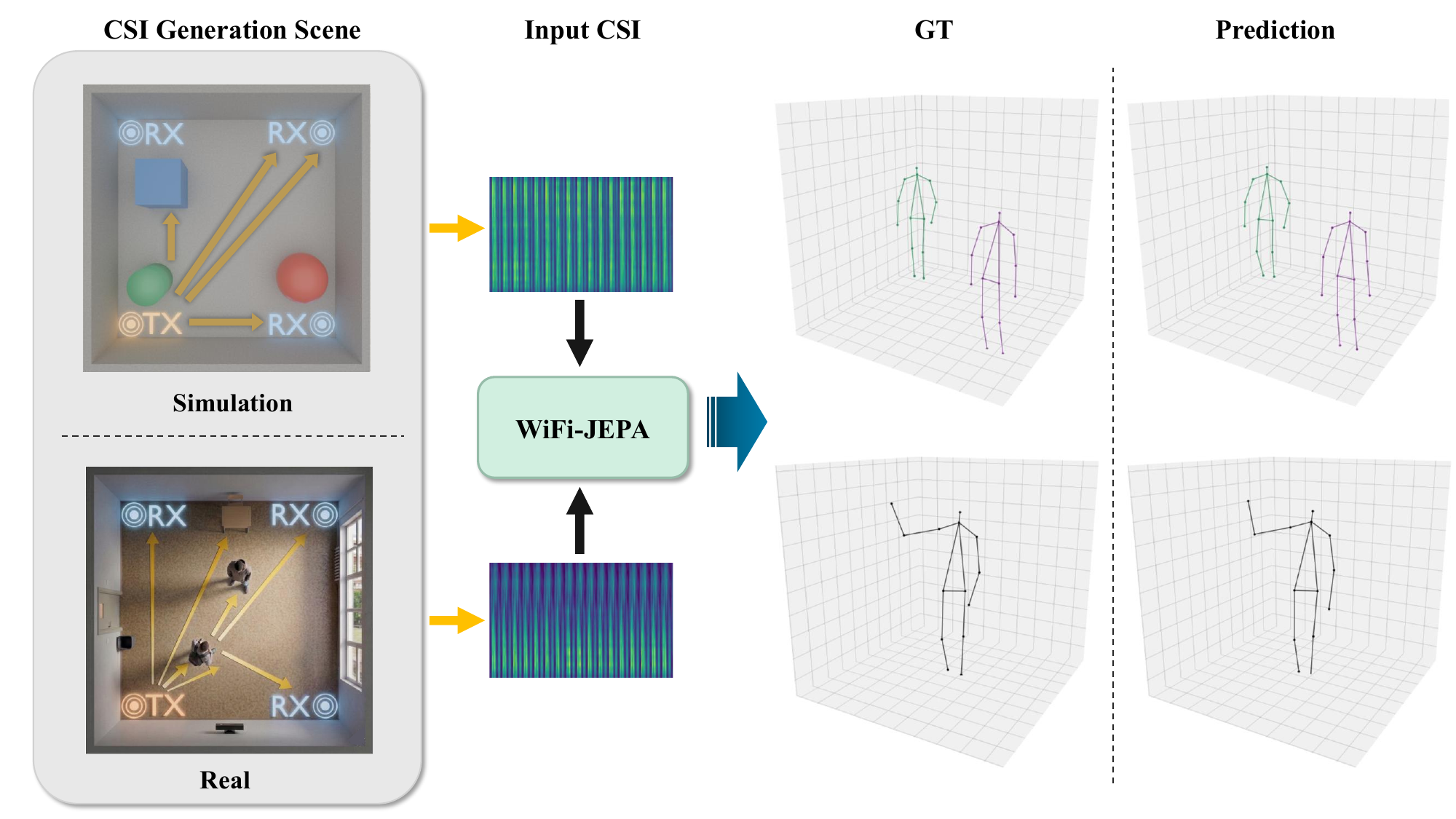}
  \caption{Overall framework of WiFi-JEPA. \emph{Left}: Pre-training data --- \simobject and real CSI from PiW3D. \emph{Center}: Generated CSI input and WiFi-JEPA. \emph{Right}: GT and predicted 3D poses.}
  \label{fig:fig1}
\end{figure}

\section{Introduction}
\label{sec:intro}

Human pose estimation (HPE) is a fundamental component of systems for health monitoring, human--computer interaction, and safety-critical perception. Despite rapid progress in camera-based HPE under line-of-sight conditions~\cite{sun2019deep,xu2022vitpose,shi2022end}, vision is often limited in camera-denied scenarios: (i) \emph{occlusion} by walls or obstacles, (ii) \emph{low or no illumination}, and (iii) \emph{privacy and regulatory} constraints that prohibit visual capture in sensitive spaces. 

\textbf{WiFi channel state information (CSI)} provides a compelling alternative. WiFi signals are ubiquitous indoors and can propagate through many common obstructions. Human motion modulates multipath propagation, which is captured as CSI---a factored, complex-valued measurement over \emph{subcarriers} (frequency), \emph{time}, and multiple \emph{Tx--Rx antenna links}. Prior work has shown that WiFi sensing can recover human-centric outputs without visual imagery, including fine-grained person perception and pose-related estimates~\cite{wang2019personinwifi,gopose,yan2024person}. Most notably, PiW3D~\cite{yan2024person} demonstrates the feasibility of multi-person 3D pose estimation with commodity WiFi even under visual occlusion, reporting a $\sim$90K-frame dataset collected in multiple indoor areas and 3D joint localization errors on the order of $\sim$100\,mm.

However, robust WiFi-based HPE still faces three bottlenecks.
First, \emph{cross-domain generalization} is fragile: performance can degrade when deployment conditions differ from training, such as changes in transceiver placement, surrounding objects and furniture.
Second, \emph{label scalability} is limited: CSI 3D pose datasets typically rely on camera-based annotation pipelines and are collected in a small number of rooms with fixed hardware~\cite{yan2024person}.
Third, CSI is \emph{noisy and hardware-dependent}. A common approach is to reshape CSI into image-like 2D grids to leverage ViT/CNN backbones; this axis-mixing can conflate the physical semantics of subcarriers, time, and links and encourage reliance on device-specific artifacts over pose-relevant dynamics. 

These challenges motivate CSI-native representation learning that (i) is robust to environment and hardware shifts, (ii) scales without dense camera-derived labels, and (iii) preserves CSI structure while exploiting multi-link spatial diversity.

We propose \textbf{WiFi-JEPA}, a self-supervised learning (SSL) framework that learns transferable CSI representations without reconstructing raw CSI. WiFi-JEPA uses a \emph{CSI-specific tokenization} and predicts masked latent representations. We further introduce \emph{link masking}, which masks entire Tx-Rx link observations and forces prediction from the remaining links, explicitly leveraging the multi-link spatial views intrinsic to CSI and central to pose recoverability.
To reduce dependence on scarce labels, we also propose a \textbf{CSI simulation pipeline} via ray-tracing (NVIDIA Sionna~\cite{hoydis2023sionna}).

Our main contributions are:
\begin{itemize}
  \item \textbf{WiFi-JEPA}: a CSI-native SSL framework with \emph{CSI-specific tokenization} and \emph{link masking} that learns CSI representations by predicting masked latent embeddings, improving \emph{cross-domain robustness}.
  \item \textbf{CSI simulation pipeline} (\simobject): a ray-tracing-based pre-training \linebreak paradigm providing evidence that dynamics diversity may matter more than geometric realism for transfer; geometric primitives outperform human meshes (\simhuman, 100.1 vs.\ 110.3\,mm MPJPE) and ${\sim}$90K simulated frames match ${\sim}$90K real frames in independent pre-training value.
  \item \textbf{State-of-the-art WiFi-based results}: combining real and simulated pre-training
   achieves 76.8\,mm single-person and 93.5\,mm multi-person MPJPE on PiW3D, improving over all prior WiFi HPE baselines under the same evaluation protocol~\cite{yan2024person,chen2025dtpose,zhou2023metafi++}.
\end{itemize}

% ===============================================================
% 2. RELATED WORK
% ===============================================================

\section{Related Work}
\subsection{Self-supervised representation learning}
\label{sec:rw:ssl}
Self-supervised learning has evolved through three paradigms: contrastive methods (SimCLR~\cite{chen2020simple}, MoCo\,v3~\cite{chen2021empirical}), momentum-teacher approaches (BYOL~\cite{grill2020bootstrap}, DINO~\cite{caron2021emerging}), and masked prediction.
MAE~\cite{he2022masked} reconstructs masked inputs in pixel space, retaining any low-level noise, whereas JEPA~\cite{assran2023self} predicts targets in a learned latent space that need not preserve such artifacts.
This latent objective is particularly relevant to WiFi CSI, where raw measurements carry hardware-specific distortions (clock offsets, quantization noise) irrelevant to downstream sensing.
Structured masking has proven effective beyond images: V-JEPA~\cite{bardes2024vjepa,assran2025vjepa2} showed that structured spatio-temporal block masking outperforms random masking for learning temporal structure in video.

Several recent works apply SSL to WiFi Channel State Information~(CSI).
SSLCSI~\cite{xu2025sslcsi} systematically benchmarked four categories of SSL algorithms for CSI-based activity recognition.
CIG-MAE~\cite{liu2025cigmae} learns to allocate masking based on per-patch information density but does not differentiate masking strategies across distinct physical axes of CSI.
AM-FM~\cite{zhu2026amfm} scales WiFi pre-training to 9.2\,M samples across nine downstream tasks including activity recognition, gesture recognition, and WiFi imaging; however, none addresses multi-person 3D skeletal pose estimation.
In the broader wireless domain, WirelessJEPA~\cite{chu2026wirelessjepa} applies JEPA to raw multi-antenna in-phase/quadrature streams for communication and RF classification; its inputs and objectives differ fundamentally from estimated CSI for human sensing.
Many CSI-SSL methods adapt image-domain augmentation or masking schemes without explicitly modeling the distinct physical semantics of CSI's frequency, spatial, and temporal axes.

\subsection{WiFi-based Human Pose Estimation}
\label{sec:rw:wifi}
Human motion modulates WiFi multipath propagation, enabling through-wall sensing without cameras.
WiPose~\cite{jiang2020towards} first demonstrated single-person 3D WiFi pose estimation with a CNN--LSTM architecture.
PiW3D~\cite{yan2024person} introduced the first multi-person 3D benchmark with a Transformer-based pose decoder.
\linebreak MetaFi++~\cite{zhou2023metafi++} processed each receiver through a shared CNN and fused features via Transformer self-attention.
HPE-Li~\cite{gian2024hpe} used dual selective-kernel CNNs with teacher--student distillation.
These supervised methods all require synchronized visual supervision (motion capture or camera-based pose annotations), limiting data scalability.

To our knowledge, DT-Pose~\cite{chen2025dtpose} is the only prior SSL method for WiFi pose estimation.
It combines MAE pretraining with temporal contrastive learning and uniformity regularization, and uses a GCN--Transformer decoder that enforces skeleton topology constraints.
Its tokenization flattens CSI into a 2D grid with fixed sinusoidal positional encoding, jointly encoding all three axes into a single sequence without preserving their independence.
Because DT-Pose further employs a decoder with explicit skeletal-topology constraints (an inductive bias absent from our PETR-based decoder), we control decoder architecture and isolate the effect of the pre-training objective in Sec.~\ref{sec:masking_analysis}.

\subsection{Simulation data for representation learning}
\label{sec:rw:sim}
Domain randomization~\cite{tobin2017domain} showed that training on diverse synthetic data with randomized scene parameters enables sim-to-real transfer without photorealistic rendering.
FractalDB~\cite{kataoka2020pre} and Dead Leaves~\cite{baradad2021learning} extended this idea to representation learning, showing that vision backbones pretrained on procedurally generated images can learn transferable features, confirming that structural diversity matters more than geometric realism.
This principle is particularly relevant for CSI, where multipath fading causes even simple moving scatterers to produce rich channel variations that encode spatial dynamics---making the diversity of motion patterns more informative than the geometric fidelity of the scatterer itself.
In the wireless domain, ray-tracing tools such as Sionna~\cite{hoydis2023sionna} and synthetic channel frameworks like DeepMIMO~\cite{alkhateeb2019deepmimo} can generate physically grounded channel data and have been used for supervised tasks (channel estimation, beam prediction), but not for self-supervised pretraining of sensing models.
Across these three threads, no prior work combines axis-aware masking with latent-space prediction tailored to the physical structure of CSI, nor has ray-tracing simulation been used for self-supervised pretraining in WiFi sensing.

% ===============================================================
% 3. Simulated CSI FROM ABSTRACT SHAPES
% ===============================================================
\section{Simulated CSI from Geometric Primitives} \label{sec:sim}

Collecting large-scale WiFi CSI datasets is expensive, requiring dedicated indoor environments, synchronized receivers, and co-located motion capture for labeling. We bypass this bottleneck with a simulation pipeline that generates benchmark-compatible CSI from scenes of simple geometric shapes---no human models or external motion data required. We refer to this geometric-primitive dataset as \simobject throughout; a human-mesh variant (\simhuman) is evaluated as a comparison in Sec.~\ref{sec:sim_analysis}.

\begin{figure}
\centering
\includegraphics[trim=0cm 3cm 5cm 0cm, width=0.8\linewidth]{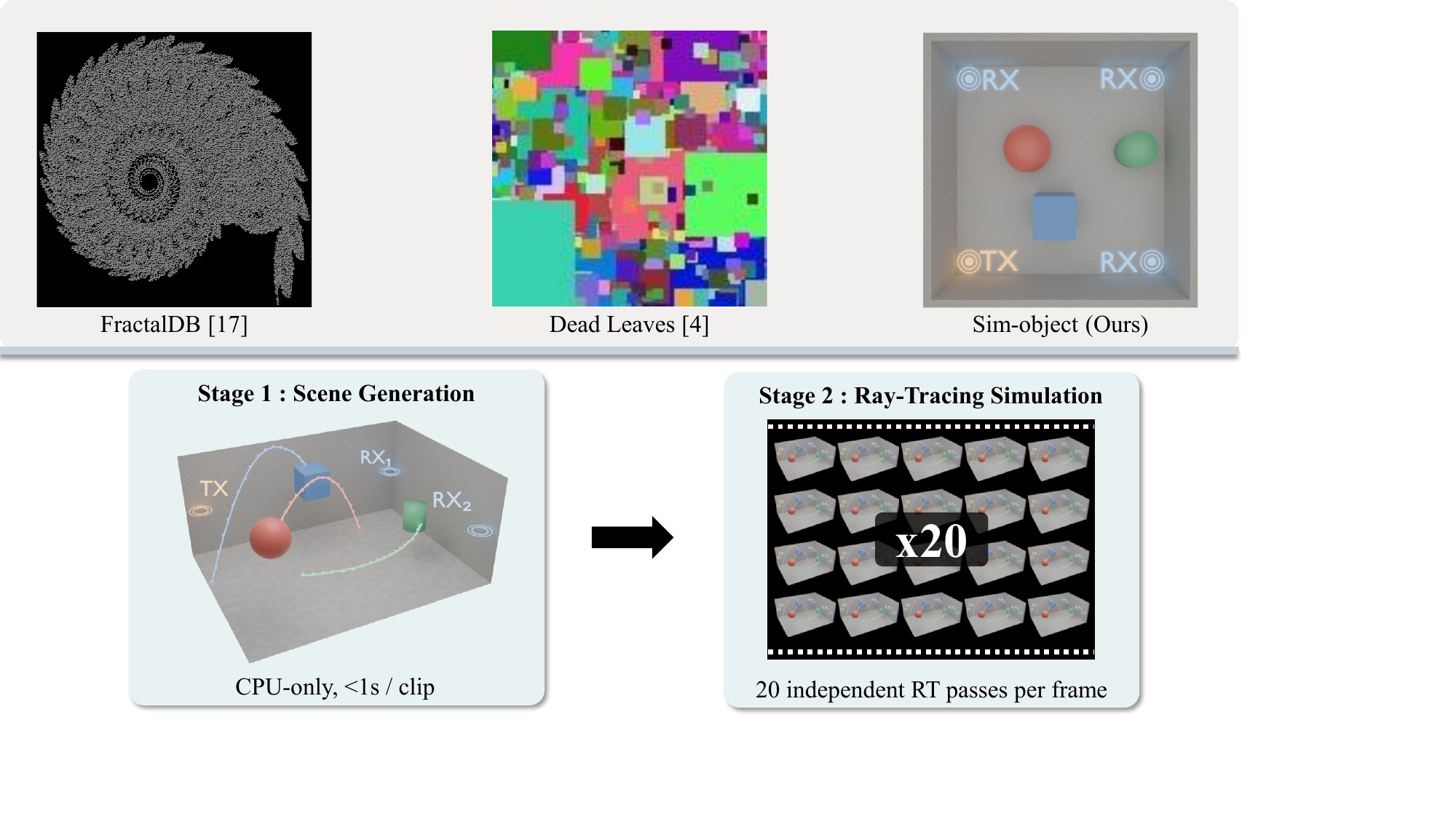}
\caption{\simobject pipeline. \textbf{Top:} Analogy to FractalDB and Dead Leaves---geometric primitives replace human models. \textbf{Bottom:} Stage~1 generates randomized scenes (CPU-only); Stage~2 runs Sionna RT with $\times$20 independent passes per frame.}
\label{fig:pipeline}
\end{figure}

Inspired by the success of non-semantic pre-training in vision (Sec.~\ref{sec:rw:sim}), we randomize room geometry, object count, trajectory, and wall materials to cover a wide range of channel conditions.

\noindent\textbf{Hypothesis.} We hypothesize that SSL pre-training may not require CSI from real human poses; rather, it needs exposure to \emph{how} WiFi signals vary across time, frequency, and space. At 5.64\,GHz (wavelength ${\approx}\,$5\,cm), a sphere and a human body differ substantially in scattering behavior---the body is articulated, non-convex, and has complex dielectric properties. Nevertheless, SSL pre-training may still succeed with geometric primitives, because the learning objective benefits from diverse spatio-temporal channel variation, rather than faithful reproduction of body-specific multipath. Specifically, we test whether \emph{dynamics diversity}---the range of positions, velocities, and directions across training scenes---matters more than geometric fidelity in Sec.~\ref{sec:sim_analysis}.

\paragraph{Stage 1: Scene generation.}
A pure-Python generator (CPU-only, ${<}1$\,s/clip) creates randomized indoor scenes. Room dimensions are sampled uniformly ($3$--$8$\,m per side, $2.5$--$4$\,m height), with walls assigned ITU-standard materials (concrete, plasterboard, wood, glass). Each scene contains 1--4 geometric primitives (sphere, cube, cylinder, ellipsoid; radius $0.1$--$0.5$\,m) that follow physics-based trajectories: initial velocities are sampled up to $3$\,m/s---exceeding typical indoor speeds per~\cite{tobin2017domain}---with elastic wall reflections producing varied spatial coverage. A single transmitter and three receivers are placed at randomized wall positions matching the SIMO configuration of the target benchmark (1\,Tx, $3 \!\times\! 3$\,Rx antennas = 9 links). The real PiW3D training set contains ${\sim}$90K frames; our 90K simulated frames thus constitute a comparable pre-training corpus.

\paragraph{Stage 2: Ray-tracing simulation.}
NVIDIA Sionna RT~\cite{hoydis2023sionna} simulates radio propagation in each scene, configured to match the target benchmark. A critical design choice is performing \textbf{20 independent} ray-tracing passes per frame: each pass computes the channel for a static scene snapshot, so we sample object positions at 20 sub-frame locations within each measurement window (${\sim}$50\,ms). Without this, a single pass yields a near-constant time axis, rendering temporal masking ineffective; 20 passes produce realistic temporal variation matching real data. The output $H \!\in\! \mathbb{C}^{20 \times 30 \times 3 \times 3}$ (time $\times$ 30 subcarrier groups $\times$ 3 receivers $\times$ 3 antennas per Rx) is decomposed into amplitude and phase, yielding a real-valued tensor of shape $(60, 20, 9)$: $60 {=} 2 {\times} 30$ subcarrier channels, $20$ time samples, and $9 {=} 3\,\text{Rx} {\times} 3\,\text{ant}$ links---analogous to a multi-channel spectrogram---for the tokenizer (Sec.~\ref{sec:masked-csi-modeling}). Generating 90K frames takes ${\sim}$10 GPU-hours on one RTX 4090.

% =============================================================================
% Section 4: WiFi-JEPA
% =============================================================================
\section{WiFi-JEPA} \label{sec:wifi-jepa}

\subsection{CSI-specific Tokenization}
\label{sec:masked-csi-modeling}

\begin{figure}
  \centering
  \includegraphics[trim=4cm 8cm 9cm 3cm,  width=0.55\linewidth]{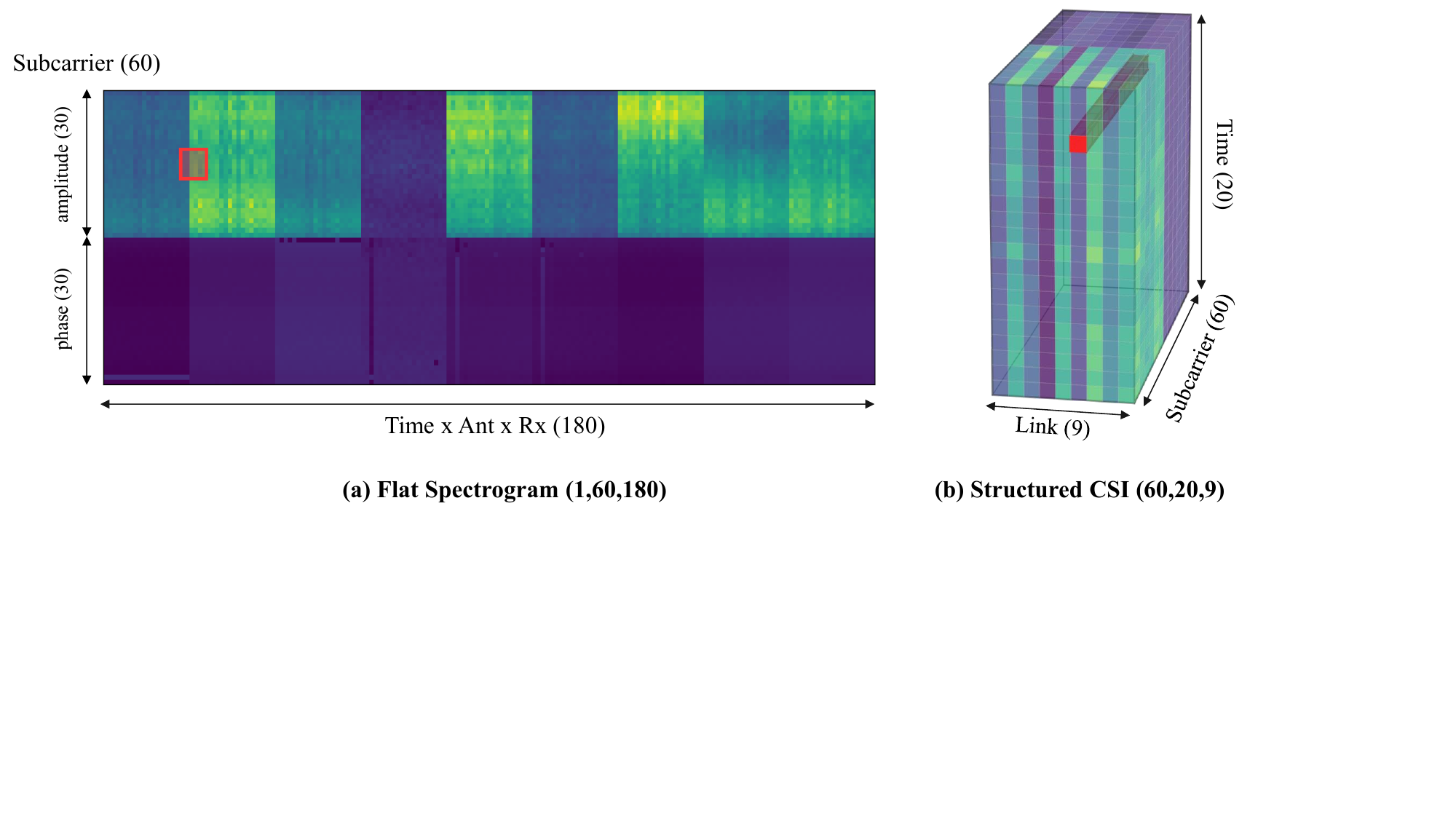}
  \caption{(a) Flattening mixes temporal and link dimensions, causing patches to cross physical boundaries. (b) Our CSI-specific tokenization keeps $(T, L)$ separate so each token corresponds to a specific spatio-temporal coordinate.}
  \label{fig:fig4}
\end{figure}

% \subsubsection{CSI-aware Tokenization.}
The raw CSI data in the PiW3D dataset~\cite{yan2024person} is a tensor of shape $N_\mathit{rx} \times N_\mathit{ant} \times T \times N_c = 3 \times 3 \times 20 \times 30$, where $N_\mathit{rx}$ is the number of receivers, $N_\mathit{ant}$ is the number of antennas per receiver, $T$ is the number of time steps, and $N_c$ is the number of subcarriers. Following~\cite{yan2024person}, we use amplitude directly and denoise phase values using PhaseFi~\cite{wang2015phasefi}, resulting in a real-valued tensor of shape $3 \times 3 \times 20 \times 60$, where $60 = 2 \times N_c$ (30 amplitude $+$ 30 denoised phase channels).

As illustrated in Fig.~\ref{fig:fig4}, PiW3D~\cite{yan2024person} flattens CSI tensor into a 2D spectrogram of shape $(60, 180)$. Although compatible with standard ViTs, this format mixes temporal sequences with spatially distinct antenna-links, blurring physical boundaries. With $6\times6$ patch embedding on the flattened grid, patches frequently cross link boundaries and the amplitude-phase boundary, merging physically disparate components within a single token.

\textbf{CSI-specific tokenization} preserves the structural integrity of the signal by reshaping the tensor into $(C, T, L) = (60, 20, 9)$, where the $L=9$ represents independent spatial links ($3R_x \times 3Ant$), each observing the same physical scene from a different spatial viewpoint. The subcarrier features ($C{=}60$) serve as the channel dimension and $T$ denotes time steps. We apply a linear embedding with patch size $(1, 1)$ over the $(T, L)$ dimensions, producing $T \times L = 180$ tokens, each representing a precise spatial-temporal coordinate. A separable 2D sinusoidal positional embedding is added.

\subsection{Link Masking} \label{sec:4.2}
In the JEPA framework, the masking strategy defines the pretext task and thus dictates the semantic quality of the acquired representations.

\begin{figure}
  \centering
  \includegraphics[trim=4cm 7cm 4cm 1cm, width=0.65\linewidth]{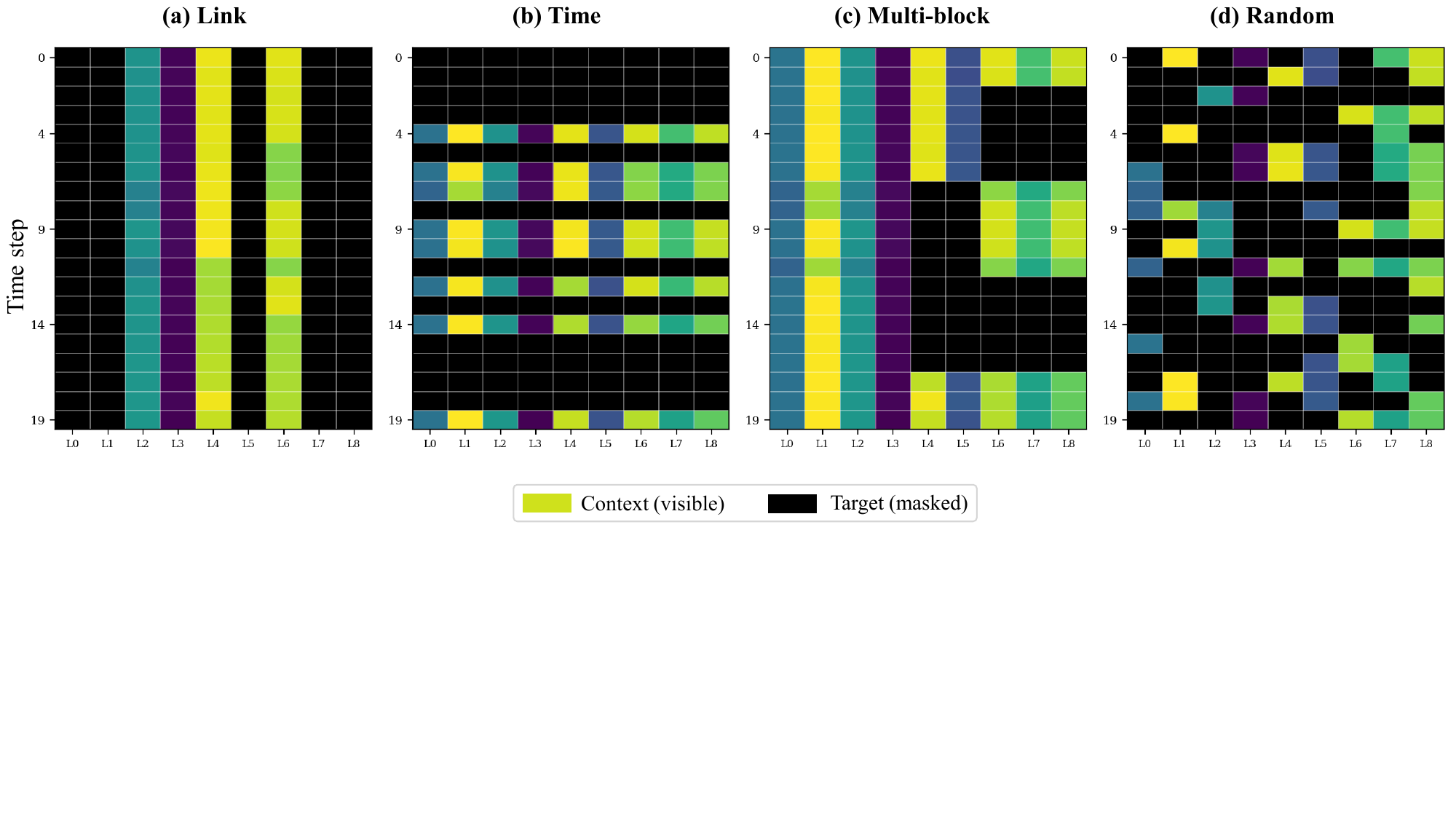}
  \caption{Four masking strategies on the $(T, L)$ token grid. The vertical axis corresponds to the 20 time steps, and the horizontal axis represents the 9 antenna links. }
  \label{fig:fig5}
\end{figure}

We propose \emph{link masking}, which masks entire columns of the $(T, L)$ token grid, as shown in Fig.~\ref{fig:fig5}(a). Given a masking ratio $r{=}0.6$, we randomly select $\mathrm{round}(r \times L) = 5$ of the 9 links as targets and use the remaining 4 links as context. All time steps of each masked link are removed simultaneously, compelling the model to predict the complete temporal signal of unseen antenna-receiver pairs from the remaining links. This aligns with the intuition of multi-view learning: as the 9 links observe the same physical scene from different viewpoints, the model must exploit cross-link spatial redundancy to reconstruct the missing signals.

To validate this hypothesis, we compare link masking against three alternatives: (i)~\emph{time masking}, which masks entire rows to test whether temporal interpolation alone suffices; (ii)~\emph{random block masking}, which masks contiguous rectangular regions of the $(T, L)$ grid following I-JEPA~\cite{assran2023self}; and (iii)~\emph{random masking}, which independently selects individual tokens regardless of grid position. Ablation results in Sec.~\ref{sec:masking_analysis} substantiate our approach: link masking consistently outperforms all alternatives, confirming that cross-link spatial correlation provides the strongest pretext signal.

\subsection{Pre-training Objective}

\begin{figure}
  \centering
  \includegraphics[trim=0 6.75cm 5cm 0, clip, width=\linewidth]{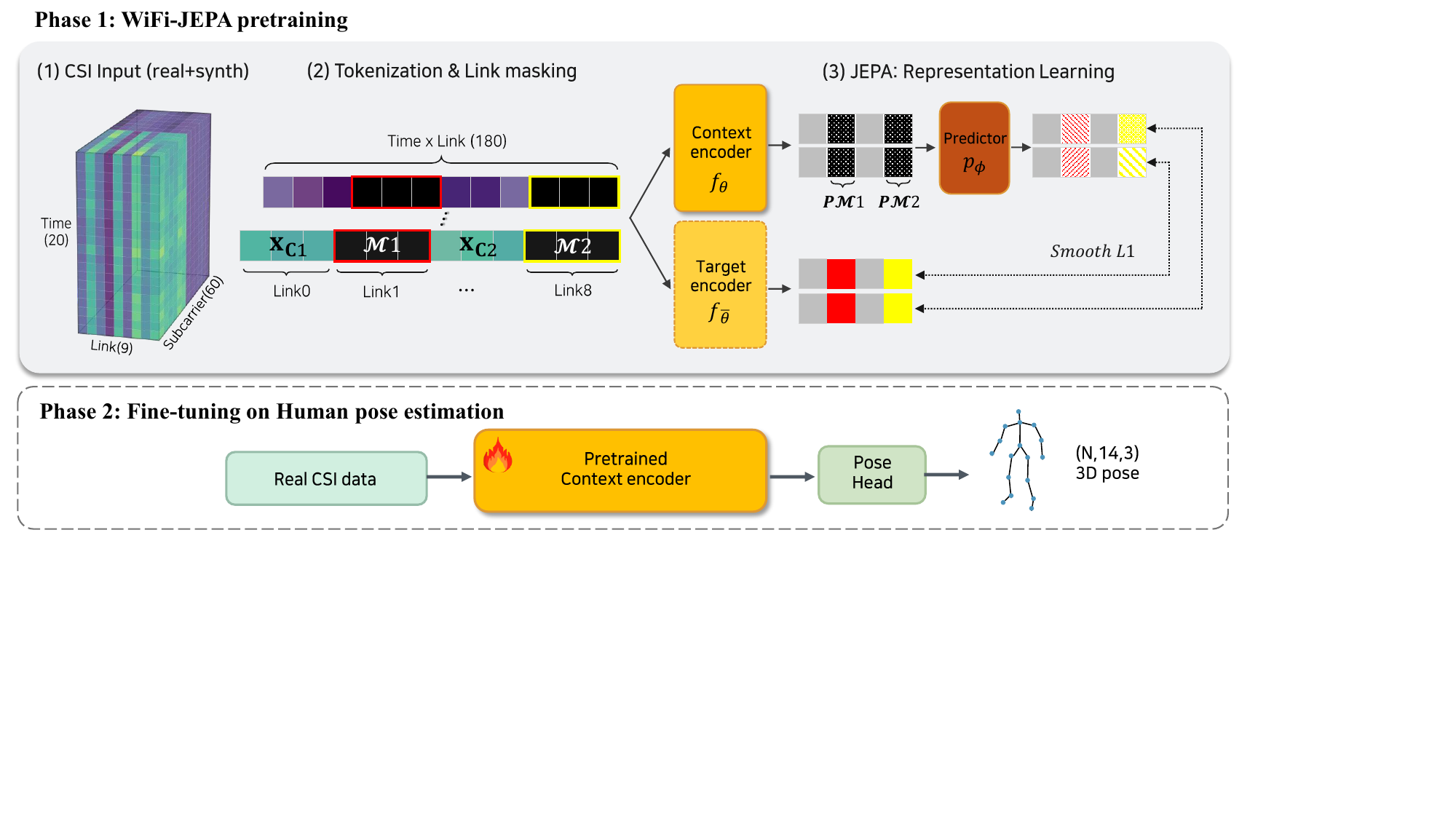}
  \caption{WiFi-JEPA architecture. \textbf{Phase~1} (top): self-supervised pre-training with link masking on the $(T,L)$ token grid. \textbf{Phase~2} (bottom): supervised fine-tuning with a PETR decoder for multi-person 3D pose estimation.}
  \label{fig:fig6}
\end{figure}

As illustrated in Phase~1 in Fig.~\ref{fig:fig6}, our architecture follows the JEPA framework~\cite{assran2023self} with three components: a context encoder $f_\theta$, a predictor $p_\phi$, and a target encoder $f_{\bar{\theta}}$. After tokenization and link masking, the context encoder takes the masked sequence as input, processing only the visible tokens $\mathbf{x}_{\mathcal{C}_i}$. The predictor receives these context embeddings concatenated with learnable mask tokens $\mathbf{p}$ at positions $\mathcal{M}_i$ and predicts the target representations. The target encoder processes all tokens, and only the representations at $\mathcal{M}$ are used as prediction targets.

We minimize the Smooth L1 loss between predicted and target embeddings:
\begin{equation}
\label{eq:loss}
\mathcal{L} = \frac{1}{|\mathcal{M}|} \sum_{i \in \mathcal{M}} \mathrm{SmoothL1}\!\left(\hat{z}_i, \; \mathrm{sg}\!\left[\mathrm{LN}(f_{\bar{\theta}}(\mathbf{x})_i)\right]\right), \quad \hat{z}_i = p_\phi\!\left(f_\theta(\mathbf{x}_{\mathcal{C}}), \; \mathbf{p}_{\mathcal{M}}\right)_i
\end{equation}
where $\mathcal{M}$ and $\mathcal{C}$ denote the masked and context token index sets, $\mathbf{p}_{\mathcal{M}}$ are the positional embeddings at masked locations, $\mathrm{sg}[\cdot]$ denotes stop-gradient, and $\mathrm{LN}$ is layer normalization applied to the target embeddings. The target encoder is updated via EMA with cosine momentum schedule from 0.996 to 1.0.

Both the context and target encoders share the same ViT backbone. We use a shallower predictor to limit its capacity and encourage the context encoder to learn more transferable, pose-relevant features. Architecture specifications and training details are provided in Sec.~\ref{sec:exp_setup}.

% -----------------------------------------------------------------------------
\subsection{Fine-tuning for 3D Pose Estimation} \label{sec:finetuning}
We integrate the pre-trained JEPA encoder into PETR~\cite{shi2022end}, with a linear projection layer mapping the encoder output from 512 to 256 dimensions to match the decoder operating dimension. During fine-tuning (Phase~2 in Fig.~\ref{fig:fig6}), we train the model end-to-end on real CSI data with differential learning rates. The pre-trained JEPA encoder is fine-tuned with $0.1\times$ the base learning rate to retain pretrained features while adapting to the task, whereas the PETR decoder and regression heads use the $1.0\times$ base learning rate.

The model predicts 14 body keypoints in 3D coordinates for $N$ detected persons ($N \times 14 \times 3$). Hungarian matching assigns queries to instances; the decoder refines predictions via
  cross-attention over encoded CSI features. Loss functions and hyperparameter settings are detailed in Sec.~\ref{sec:exp_setup}.

% ===============================================================
% 5. EXPERIMENTS
% ===============================================================
\section{Experiments}
\label{sec:experiments}

We evaluate WiFi-JEPA on the PiW3D dataset~\cite{yan2024person}, the only public WiFi-CSI benchmark that includes multi-person 3D pose annotations. Our experiments address three questions: (i)~How much does WiFi-JEPA advance the state of the art in WiFi-based 3D pose estimation? (ii)~Why JEPA over alternative SSL objectives, and do the proposed structure-aware tokenization and link masking contribute to downstream performance? (iii)~Can \simobject complement or even replace real data for pre-training?

% ============================================================
% §5.1 Experimental Setup (~0.5 page)
% ============================================================

\subsection{Experimental Setup}
\label{sec:exp_setup}

\subsubsection{Dataset.}
PiW3D~\cite{yan2024person} provides synchronized WiFi-CSI and 3D pose labels (14 joints) captured in three indoor environments---an office, a classroom, and a corridor---each measuring approximately $4\, \text{m} \times 3.5\, \text{m}$, using one transmitter and three Intel~5300 receivers (three antennas each) at 5.64\,GHz with 30 subcarriers. Seven volunteers performed eight daily actions (e.g., walking, stretching, bending over, sitting down) across these environments, yielding diverse multipath conditions. Each CSI sample has raw dimensions $1 \!\times\! 3 \!\times\! 3 \!\times\! 30 \!\times\! 20$ (\#Tx, \#Rx, \#Ant, \#Subcarrier, \#Time). The training set contains 89{,}946 frames; the test set contains 7{,}824 frames spanning single-person (2{,}586), two-person (3{,}184), and three-person (2{,}054) scenarios.

\subsubsection{Metrics.}
We report Mean Per-Joint Position Error (\textbf{MPJPE}, mm) as the primary metric---the mean Euclidean distance between predicted and ground-truth 3D joint positions. We additionally report Procrustes-aligned MPJPE (\textbf{PA-MPJPE}), which removes global translation, rotation, and scale to isolate articulated pose accuracy, and Percentage of Correct Keypoints (\textbf{PCK@$\tau$}) at $\tau{=}20,50$\,mm. For finer-grained analysis, we also report per-dimension absolute errors along the horizontal~(x), vertical~(y), and depth~(z) axes, following~\cite{yan2024person}.

\subsubsection{Implementation details.}
All training is conducted on a single NVIDIA RTX 4090 GPU. \textbf{Pre-training:} The context and target encoders are (12 layers, embedding dimension 512, 8 heads), the predictor is a ViT (8 layers, 512 dimensions, 16 heads). The target encoder is updated via EMA with momentum scheduled from 0.996 to 1.0. We pre-train for 100 epochs with AdamW (batch size 64, learning rate $5{\times}10^{-4}$, weight decay 0.05), using Smooth L1 loss with layer-normalized targets. Link masking randomly drops 60\% of antenna links per sample. Pre-training data comprises ${\sim}$90K real frames from PiW3D and ${\sim}$90K synthetic frames from \simobject (Sec.~\ref{sec:sim}); the real-only variant uses PiW3D frames alone. \textbf{Fine-tuning:} We attach the pre-trained WiFi-JEPA context encoder to a PETR-style Transformer decoder (5 layers, 256 dimensions, 8 heads) and fine-tune for 100 epochs with AdamW (base learning rate $2{\times}10^{-5}$, weight decay $10^{-4}$). The encoder learning rate is scaled to $0.1\times$ the base rate.

% ============================================================
% §5.2 Main Results (~1.0 page)
% Table 1: SOTA comparison
% Table 2: Challenging Scenarios
% ============================================================

\subsection{Main Results} \label{sec:main_results}
\subsubsection{Comparison with State-of-the-Art}
We compare against five WiFi-CSI pose estimation methods (see Sec.~\ref{sec:rw:wifi} for architectural details):
  WiPose~\cite{jiang2020towards},
  HPE-Li~\cite{gian2024hpe},
  MetaFi++~\cite{zhou2023metafi++} (single-person),
  DT-Pose~\cite{chen2025dtpose} (the only prior SSL method),
  and PiW3D~\cite{yan2024person} (the first multi-person 3D method).

\begin{table}[t]
\centering
\caption{Comparison with prior WiFi-CSI pose estimation methods on PiW3D dataset. Best in \textbf{bold}, second-best \underline{underlined}. ``--'' (gray) indicates metrics not reported in the original paper. $^\dagger$ Values taken directly from the original publications.}
\label{tab:main}
\renewcommand{\arraystretch}{1.1}

\resizebox{\textwidth}{!}{
\begin{tabular}{@{}llcccccccc@{}}
\toprule
\multirow{2}{*}{\textbf{Method}} & \multirow{2}{*}{\textbf{Venue}} &
\multicolumn{4}{c}{\textbf{Single-person}} &
\multicolumn{4}{c}{\textbf{Multi-person}} \\
\cmidrule(lr){3-6}\cmidrule(lr){7-10}
& & MPJPE$\downarrow$ & PA$\downarrow$ & PCK@20$\uparrow$ & PCK@50$\uparrow$ &
MPJPE$\downarrow$ & PA$\downarrow$ & PCK@20$\uparrow$ & PCK@50$\uparrow$ \\
\midrule
WiPose~\cite{jiang2020towards} & MobiCom'20 & $101.8^\dagger$ & \gr & \gr & \gr & \gr & \gr & \gr & \gr \\
MetaFi++~\cite{zhou2023metafi++} & IoT-J'23 & $132.0^\dagger$ & $75.8^\dagger$ & 62.0 & 89.3 & \gr & \gr & \gr & \gr \\
HPE-Li~\cite{gian2024hpe} & ECCV'24 & $120.2^\dagger$ & $69.5^\dagger$ & 59.1 & 87.5 & \gr & \gr & \gr & \gr \\
DT-Pose~\cite{chen2025dtpose} & arXiv'25 & $90.0^\dagger$ & $58.7^\dagger$ & 72.1 & 90.6 & \gr & \gr & \gr & \gr \\
PiW3D (baseline)~\cite{yan2024person} & CVPR'24 & $91.7^\dagger$ & 55.1 & 69.3 & 95.2 & $107.2^\dagger$ & \underline{65.6} & 58.1 & 91.8 \\
\midrule
WiFi-JEPA (real) & \multicolumn{1}{c}{--} & \underline{78.2} & \textbf{53.9} & \underline{74.5} & \underline{96.6} & \underline{97.1} & 67.2 & \underline{59.3} & \underline{93.0} \\
WiFi-JEPA (real+sim) & \multicolumn{1}{c}{--} & \textbf{76.8} & \underline{54.0} & \textbf{75.9} & \textbf{96.7} & \textbf{93.5} & \textbf{65.1} & \textbf{61.5} & \textbf{93.7} \\
\bottomrule
\end{tabular}
}
\end{table}

Table~\ref{tab:main} summarizes the results. We note that WiPose, MetaFi++, HPE-Li, and DT-Pose do not report multi-person (MP) results; we compare them only in the single-person (SP) scene. WiFi-JEPA (real+sim) achieves 76.8\,mm single-person MPJPE, outperforming the previous best method DT-Pose (90.0\,mm) by 13.2\,mm ($-$14.7\%) and the PiW3D baseline (91.7\,mm) by 14.9\,mm ($-$16.2\%). PCK@20 also increases from 72.1\% (DT-Pose) to 75.9\%. In multi-person scenes---where only PiW3D reports prior results---WiFi-JEPA reduces MPJPE from 107.2\,mm to 93.5\,mm ($-$12.8\%) and raises PCK@20 from 58.1\% to 61.5\%. Notably, the real-only pre-training variant already surpasses all baselines (78.2\,mm SP, 97.1\,mm MP), and adding simulation data yields a further 1.4\,mm and 3.6\,mm reduction in SP and MP respectively, demonstrating that \simobject provides complementary pre-training signal.

\begin{figure}[t]
  \centering
  \includegraphics[trim=0cm 0cm 0cm 0cm, clip, width=\linewidth]{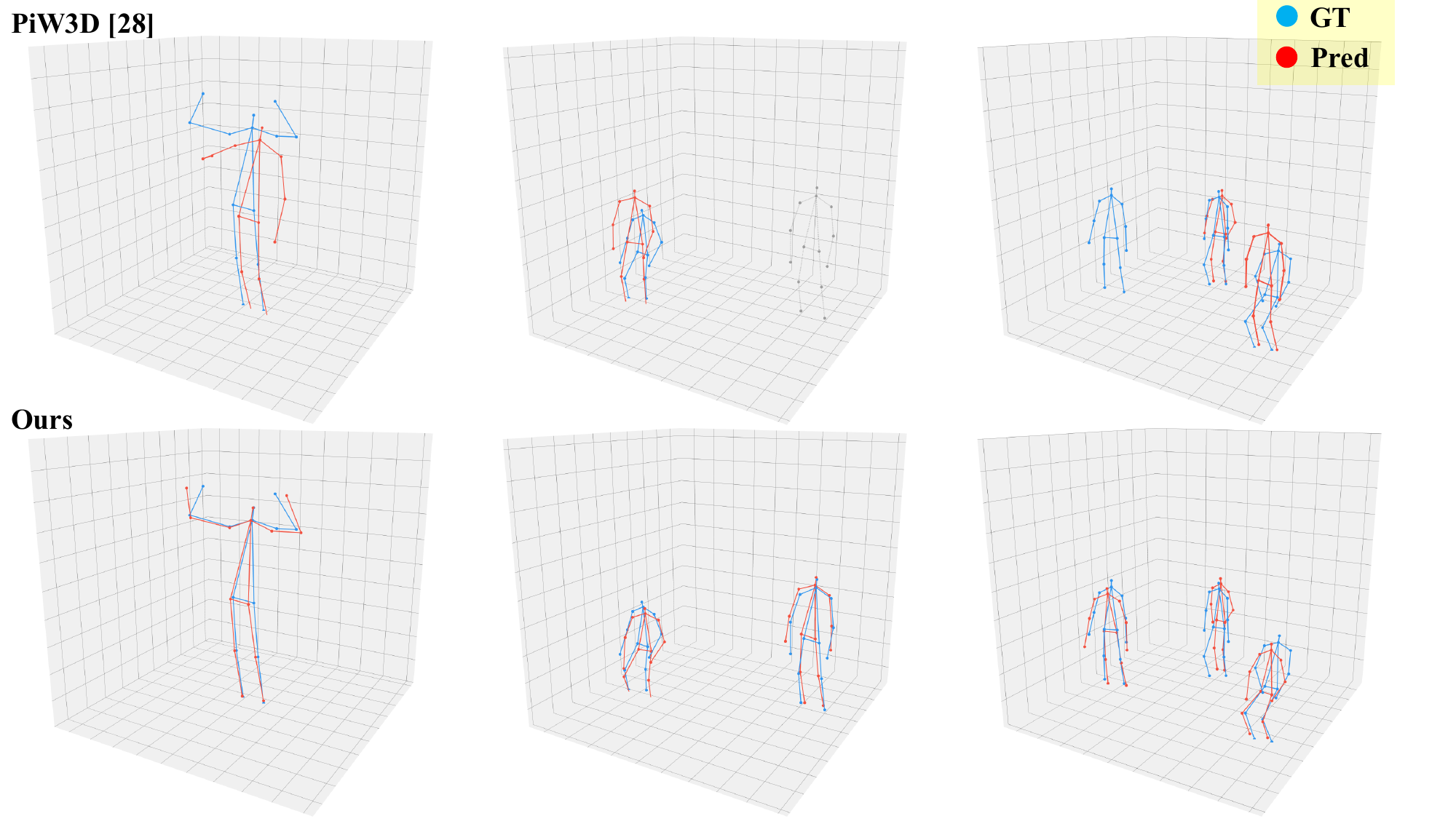}
  \caption{Qualitative comparison of 3D pose estimation. Top row: PiW3D baseline~\cite{yan2024person}. Bottom row: WiFi-JEPA (ours). From left to right: 1-person, 2-person, and 3-person scenes.}
  \label{fig:fig7}
\end{figure}

\begin{figure*}[b]
    \centering
    % --- 왼쪽: 표 ---
    \begin{minipage}[t]{0.50\linewidth}
    \centering
    \setlength{\tabcolsep}{6pt} % 자간을 살짝 넓혀 세로선 없이도 구분이 잘 가도록 함
    \renewcommand{\arraystretch}{1.3}
    
    \captionof{table}{\textbf{Leave-one-environment-out cross-domain evaluation (MPJPE, mm).}
    Off./Class./Corr.: held-out test environment (Office/Classroom/Corridor). Both pre-training
    and fine-tuning exclude the held-out environment.}
    \label{tab:loo_env}
    
    \vspace{0pt} 
    \resizebox{\linewidth}{!}{%
    \begin{tabular}{ll ccc} % <- 세로선 제거
    \toprule
    \textbf{Method} & \textbf{Metric} & \textbf{Off.} & \textbf{Class.} & \textbf{Corr.} \\
    \midrule
    \multirow{2}{*}{\shortstack{WiFi-JEPA\\(ours)}}
      & MPJPE & 248.4 & 428.2 & 296.0 \\
      & Mean & \multicolumn{3}{c}{\textbf{324.2}} \\
    \cmidrule(lr){1-5} % <- 얇은 중간 선을 넣어, 우리 모델과 베이스라인의 구역을 확실히 분리 (회색선 대체)
    \shortstack{PiW3D\\(baseline)} & Mean & \multicolumn{3}{c}{\textbf{626.4}} \\
    \bottomrule
    \end{tabular}%
    }
    \end{minipage}%
    \hfill
    % --- 오른쪽: 피규어 ---
    \begin{minipage}[t]{0.45\linewidth}
    \centering
    \vspace{0pt} 
    \includegraphics[trim=0 8cm 17cm 0, width=\linewidth]{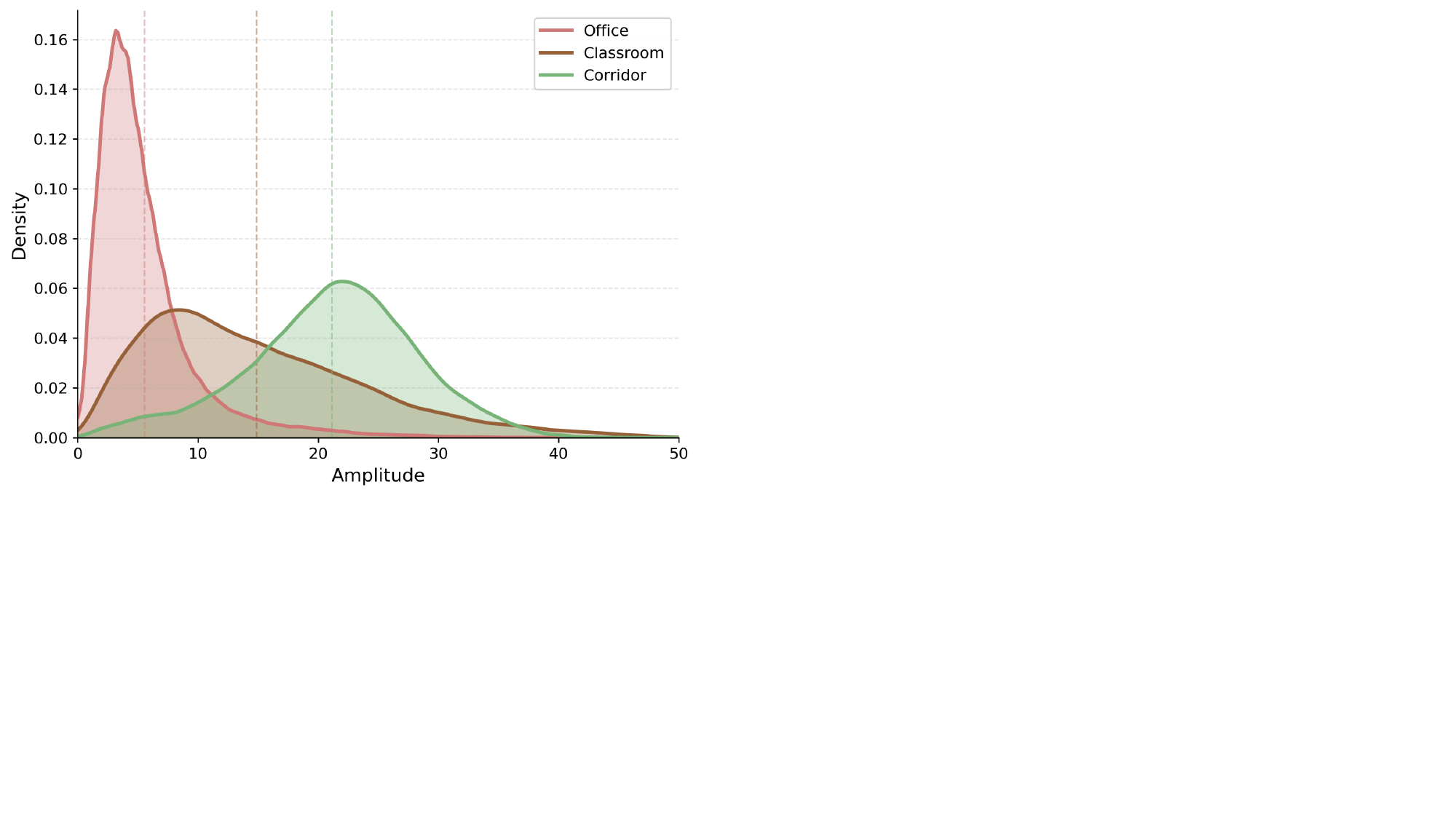}
    \captionof{figure}{Amplitude distribution shift across 3 environments.}
    \label{fig:domain_gap}
    \end{minipage}
\end{figure*}

\subsubsection{Challenging Scenarios}
\paragraph{Cross-Domain Generalization}
WiFi CSI amplitude profiles vary across environments due to multipath propagation differences. For instance, the mean amplitude on a single receiver--antenna-link ranges from 5.7 in Office to 20.7 in Corridor (Fig.~\ref{fig:domain_gap}), with similar variation across all nine links. To test whether WiFi-JEPA representations transfer across environments, we conduct a leave-one-environment-out (LOO) evaluation: both pre-training and fine-tuning use data from two environments only, and the model is tested on the held-out environment (Table~\ref{tab:loo_env}).  WiFi-JEPA achieves 248.4\,mm on Office, 428.2\,mm on Classroom, and 296.0\,mm on Corridor, yielding a mean MPJPE of 324.2\,mm---a 48.2\% reduction over the PiW3D baseline (626.4\,mm). While absolute errors remain high, WiFi-JEPA halves the baseline error. This is consistent with the masking ablation in Section~\ref{sec:masking_analysis}, where link masking consistently outperforms alternative strategies, indicating that cross-link correlations are key to environment-robust representations. Furthermore, we evaluate WiFi-JEPA against a standard supervised Domain Adaptation (DA) baseline, DANN~\cite{ganin2016domain}, in a few-shot cross-environment setup utilizing only 10\% of target environment labels. WiFi-JEPA consistently outperforms Supervised+DANN across all held-out environments, reducing the macro mean MPJPE from 208.6 mm to 171.3 mm (a 17.9\% error reduction). This demonstrates that our SSL pre-training yields substantial generalization benefits that go beyond standard adversarial DA techniques.

\begin{table*} % <- [b]에서 [t]로 변경하여 하단 배치 오류 방지
    \centering
    % === 왼쪽 표 세트 (위쪽 맞춤) ===
    \begin{minipage}[t]{0.48\linewidth}
    \centering
    \footnotesize
    
    \captionof{table}{\textbf{Per-person-count MPJPE} (mm). Performance degrades gracefully as
    the number of simultaneous persons increases.}
    \label{tab:perperson}
    
    \renewcommand{\arraystretch}{1.3}
    \begin{tabular}{llccc}
    \toprule
    Metric & Method & 1P & 2P & 3P \\
    \midrule
    \multirow{2}{*}{MPJPE$\downarrow$}
     & PiW3D & 91.7  & 108.1 & 125.3 \\
     & Ours & \textbf{76.8} & \textbf{96.0} & \textbf{110.7} \\
    \bottomrule
    \end{tabular}
    \end{minipage}%
    \hfill
    % === 오른쪽 표 세트 (위쪽 맞춤) ===
    \begin{minipage}[t]{0.50\linewidth}
    \centering
    \footnotesize
    
    \captionof{table}{\textbf{Extremity error breakdown.} Mean L/R elbow and hand errors with
    directional decomposition (mm). PiW3D: Person-in-WiFi-3D baseline~\cite{yan2024person}.}
    \label{tab:extremity}

    \renewcommand{\arraystretch}{1.05}
    \begin{tabular}{l cc cc}
    \toprule
    \multirow{2}{*}{Metric} & \multicolumn{2}{c}{Elbows} & \multicolumn{2}{c}{Hands} \\
    \cmidrule(lr){2-3}\cmidrule(lr){4-5}
     & PiW3D & Ours & PiW3D & Ours \\
    \midrule
    MPJPE$\downarrow$  & 128.5 & \textbf{60.6}  & 192.5 & \textbf{68.7}  \\
    Horiz$\downarrow$  & 68.2  & \textbf{31.1}  & 94.4  & \textbf{34.4}  \\
    Vert$\downarrow$   & 56.8  & \textbf{35.0}  & 104.8 & \textbf{40.0}  \\
    Depth$\downarrow$  & 73.9  & \textbf{23.2}  & 92.9  & \textbf{26.4}  \\
    \midrule
    Improv. & \multicolumn{2}{c}{\textbf{\color{red}$\downarrow$52.8\%}} & % <- \color{red} 완벽 제거
    \multicolumn{2}{c}{\textbf{\color{red}$\downarrow$64.3\%}} \\ % <- \color{red} 완벽 제거
    \bottomrule
    \end{tabular}
    \end{minipage}
\end{table*}

\paragraph{Multi-Person Scenes and Per-joint Analysis}
A known limitation of WiFi-CSI pose estimation is performance degradation in multi-person scenarios, where overlapping signals make it difficult to disentangle individual poses. WiFi-JEPA improves across all person
  counts: single-person MPJPE drops from 91.7\,mm to 76.8\,mm ($-$16.2\%), two-person from 108.1\,mm to 96.0\,mm ($-$11.2\%), and three-person from 125.3\,mm to 110.7\,mm ($-$11.7\%) (Table~\ref{tab:perperson}).
  
The PiW3D baseline struggles most with elbows and hands (160.5\,mm), because arms exhibit faster, more varied motion and occupy a smaller spatial cross-section~\cite{yan2024person}. WiFi-JEPA reduces this to 64.7\,mm, a 59.7\% improvement (Table~\ref{tab:extremity}), with depth error dropping from 92.9\,mm to 26.4\,mm for hands.

%=====================================================================================================
\subsection{Effect of Simulated CSI}
\label{sec:sim_analysis}

We now examine the hypothesis from Sec.~\ref{sec:sim}. Table~\ref{tab:sim} presents a dataset-level comparison: all rows use the same WiFi-JEPA encoder, PETR decoder, and fine-tuning protocol; only the pre-training data varies. We additionally compare against \simhuman, a human-mesh variant that uses Blender-animated characters with motion-captured animation clips instead of geometric primitives. Both simulated datasets use the same number of clips and are sampled to 90K frames, so the comparison isolates the effect of scene content: geometric primitives with randomized physics trajectories (\simobject) versus animated human meshes with motion-capture sequences (\simhuman).

\begin{table}
  \centering
  \caption{Effect of pre-training data on downstream MPJPE (mm). All rows share the same encoder, decoder, and fine-tuning protocol. $\Delta$: improvement over no pre-training.}
  \label{tab:sim}
  \small
  \begin{tabular}{lcc cc}
  \toprule
  \multirow{2}{*}{Pre-train data} & \multicolumn{2}{c}{Frames} &
  \multirow{2}{*}{MPJPE$\downarrow$} & \multirow{2}{*}{$\Delta$} \\
  \cmidrule(lr){2-3}
   & Real & Sim & & \\
  \midrule
  None (scratch) & --- & --- & 102.4 & --- \\
  \midrule
  \simobject & --- & 90K & 100.1 & $-$2.3 \\
  \simhuman & --- & 90K & 110.3 & $+$7.9 \\
  \midrule
  Real only & 90K & --- & 97.1 & $-$5.3 \\
  Real + \simobject & 45K & 45K & 97.2 & $-$5.2 \\
  Real + \simobject & 90K & 90K & \textbf{93.5} & \textbf{$-$8.9} \\
  \bottomrule
  \end{tabular}
  \end{table}

\subsubsection{\simobject suffices for pre-training.}
Pre-training on \simobject (100.1\,mm) outperforms \simhuman (110.3\,mm), with the latter causing \emph{negative transfer} ($+$7.9\,mm worse than no pre-training), likely because the fixed motion-capture sequences lack the trajectory variation needed to learn generalizable channel features. Because \simhuman uses anatomically realistic scatterers yet performs worse, the result directly supports the hypothesis that trajectory diversity---randomized physics with varied velocities and elastic reflections---is more important than scatterer realism for CSI pre-training. 
Furthermore, \simobject alone (100.1\,mm) is close to real-data-only pre-training (97.1\,mm), with a gap of only 3.0\,mm (2.3 vs.\ 5.3\,mm reduction from scratch). That is, ${\sim}$90K simulated frames provide comparable pre-training value to ${\sim}$90K real frames.

\subsubsection{Complementarity of Simulated and Real Data.}
Combining 90K simulated and 90K real frames reduces MPJPE by 8.9 mm over training from scratch, compared with a 5.3 mm reduction from real-only pre-training. This suggests that the simulation pipeline provides channel variation patterns absent from the real dataset, such as diverse room geometries and wall materials. The simulated data is generated at a cost of ${\sim}$10 GPU-hours on one RTX 4090 (Sec.~\ref{sec:sim}), while the real frames simply reuse the existing training set as unlabeled data.

% ============================================================
% §5.4 Masking Strategy Analysis (~0.5 page)
% Table 5: Masking strategy
% ============================================================
\subsection{Effect of WiFi-JEPA Pre-training} \label{sec:masking_analysis}
\subsubsection{Comparison with SSL Baselines}

\begin{table}
    \centering
    \caption{All methods use the same ViT backbone and PETR-style
    decoder, pre-trained on ${\sim}$90K real frames for 100 epochs, then fine-tuned
  identically. Best in \textbf{bold}, second-best \underline{underlined}.}
    \label{tab:ssl}
    \small
    \begin{tabular}{llcccc}
    \toprule
    Method & SSL Type & MPJPE$\downarrow$ & PA$\downarrow$ & PCK@20$\uparrow$ &
  PCK@50$\uparrow$ \\
    \midrule
    SimMIM~\cite{xie2022simmim} & MIM & 145.6 & 84.0 & 41.5\% & 84.3\% \\
    MAE~\cite{he2022masked} & MIM & 130.3 & 77.4 & 47.6\% & 87.4\% \\
    BYOL~\cite{grill2020bootstrap} & Self-distill & 144.4 & 83.2 & 42.5\% & 84.1\% \\
    MoCoV3~\cite{chen2021empirical} & Contrastive & 141.5 & 82.6 & 43.4\% & 84.9\% \\
    \midrule
    WiFi-JEPA (w/o pretrain) & -- & \underline{102.4} & \textbf{65.1} & \underline{57.8\%} & \underline{91.8\%}\\
    WiFi-JEPA (w/ pretrain) & JEPA & \textbf{97.1} & \underline{67.2} & \textbf{59.3\%} &
  \textbf{93.0\%} \\
    \bottomrule
    \end{tabular}
  \end{table}

WiFi-JEPA outperforms all baselines in MPJPE and PCK by a wide margin (Table~\ref{tab:ssl}). The gap to the
  strongest competitor, MAE, is 33.2\,mm MPJPE ($130.3 \to 97.1$) and $+$11.7\,pp in PCK@20
  ($47.6\% \to 59.3\%$). PA-MPJPE remains comparable to the no-pre-training baseline,
  indicating that gains concentrate in global localization rather than relative joint
  proportions. All four baselines actually \emph{degrade} performance relative to no
  pre-training (102.4\,mm). MIM methods  must reproduce raw CSI values including
  hardware artifacts, which may bias the encoder toward device-specific patterns.
  Self-distillation and contrastive methods require augmentations that preserve
   channel structure, however standard image augmentations lack physical grounding for CSI. These
  negative results may partly reflect the difficulty of adapting vision-native methods to a
  new modality. WiFi-JEPA avoids these pitfalls by predicting in a learned latent space and
  exploiting the factored $(C, T, L)$ structure of CSI.

% =====================================================================================================
\begin{table}[b]
\centering

\begin{minipage}[t]{0.56\linewidth}
    \caption{\textbf{Masking strategy ablation.} All variants use the same backbone and training schedule, pre-trained on ${\sim}$90K real frames for 100 epochs.}
    \label{tab:ablation_mask}
    \centering
    \small
    \begin{tabular}{@{}lccc@{}}
    \toprule
    Method & MPJPE$\downarrow$ & PA-MPJPE$\downarrow$ & PCK@50$\uparrow$ \\
    \midrule
    Random & 105.14 & 69.30 & 88.48\% \\
    Multi-block & 102.06 & 67.71 & 90.54\% \\
    Time & 104.24 & 68.25 & 88.22\% \\
    \textbf{Link} & \textbf{97.10} & \textbf{67.20} & \textbf{93.00\%} \\
    \bottomrule
    \end{tabular}
\end{minipage}%
\hfill
\begin{minipage}[t]{0.40\linewidth}
    \caption{\textbf{Effect of structure-aware tokenization.} w/o: 2D spectrogram (1$\times$60$\times$180); w/: CSI-specific tokenization (60$\times$20$\times$9).}
    \label{tab:ablation_tok}
    \centering
    \small
    \begin{tabular}{@{}lcc@{}}
    \toprule
    Metric & w/o csi-tok. & w/csi-tok.\\
    \midrule
    PA-MPJPE$\downarrow$ & 70.98 & \textbf{67.20} \\
    MPJPE$\downarrow$ & 111.85 & \textbf{97.10} \\
    \bottomrule
    \end{tabular}
\end{minipage}

\end{table}

\subsubsection{Ablation Studies}
\emph{Masking strategies:} We compare four masking strategies described in
  Sec.~\ref{sec:4.2}: random (individual patches),
  multi-block~\cite{assran2023self} (contiguous rectangles), temporal
  (entire time rows), and link (entire receiver--antenna columns). All
  variants are pre-trained on ${\sim}$90K real frames for 100 epochs with
  the same hyperparameters, then fine-tuned (Table~\ref{tab:ablation_mask}). Link masking achieves the lowest MPJPE
  (97.1\,mm) and highest PCK@50 (93.0\%), outperforming random masking
  by 8.04\,mm and 4.52\,pp respectively. \emph{CSI-specific tokenization:} Table~\ref{tab:ablation_tok} compares our CSI-specific tokenization against a 2D spectrogram baseline that flattens the CSI tensor into a single-channel $60\times180$ image with $6\times6$ patches. Both models use the same WiFi-JEPA and link masking.
  CSI-specific tokenization reduces MPJPE by 14.75\,mm (111.85$\to$97.1) and PA-MPJPE by
  3.78\,mm, confirming that preserving the physical axis semantics of subcarrier, time, and
  link enables more effective representation learning.

% ===============================================================
% 6. CONCLUSION
% ===============================================================
\section{Conclusion} \label{sec:conclusion}

We presented WiFi-JEPA, a self-supervised framework for WiFi-CSI-based 3D human pose estimation that predicts masked latent embeddings on the factored $(C,T,L)$ CSI tensor. Structure-aware tokenization and link masking capture cross-link spatial correlations central to body localization, while ray-tracing simulation provides scalable pre-training data without annotation. Across single- and multi-person 3D pose estimation on PiW3D, WiFi-JEPA sets a new state of the art and nearly halves cross-environment error, and it consistently improves over training from scratch, whereas four vision-native SSL objectives degrade it. Notably, simulated CSI from simple moving primitives offers pre-training value comparable to real data, indicating that the diversity of channel dynamics matters more than the geometric realism of the scatterer.

\noindent\textbf{Limitations.}
WiFi-JEPA improves global joint localization more than relative joint configuration (PA-MPJPE stays close to the from-scratch baseline). Cross-environment error, though substantially reduced, remains far higher than same-environment performance. Our evaluation is confined to a single dataset with fixed hardware, so generalization across antenna configurations and frequency bands remains open.

\noindent More broadly, CSI-native self-supervised learning, which respects the physical structure of the wireless channel rather than treating CSI as a generic image, is a promising direction for WiFi sensing beyond pose estimation.

\subsubsection*{Acknowledgements}
This work was supported by the G-LAMP Program of the National Research Foundation of Korea (NRF) grant funded by the Ministry of Education (No. RS-2025-25441317); the Ministry of Science and ICT (MSIT) and the National IT Industry Promotion Agency (NIPA) through the Advanced GPU Utilization Support Program (02-26-01-0499); the NRF grant funded by the MSIT (RS-2025-16071992); the Korea Institute for Advancement of Technology (KIAT) grant funded by the Korea government(MOTIE) (RS-2026-25530975, HRD Program for Industrial Innovation); the MSIT under the Convergence security core talent training business support program(IITP-2024 2024-RS-2024-00426853) supervised by the IITP(Institute of Information \& Communications Technology Planning \& Evaluation); and the IITP grant funded by the MSIT (No. 2022-0-00986, Development of artificial intelligence-based base station electromagnetic wave human exposure prediction algorithm).

% ---- Bibliography ----
%
% BibTeX users should specify bibliography style 'splncs04'.
% References will then be sorted and formatted in the correct style.
%
\bibliographystyle{splncs04}
\bibliography{main}
\end{document}